\documentclass[letterpaper, 10 pt, conference]{ieeeconf}
\usepackage{float}
\IEEEoverridecommandlockouts                             
\usepackage{hyperref}
\usepackage{comment}
\usepackage{graphics} % for pdf, bitmapped graphics files
\usepackage{epsfig} % for postscript graphics files
\usepackage{mathptmx} % assumes new font selection scheme installed
\usepackage{times} % assumes new font selection scheme installed
\usepackage{amsmath} % assumes amsmath package installed
\usepackage{amssymb}  % assumes amsmath package installed
\let\labelindent\relax
\usepackage{enumitem}
\usepackage{graphicx,subcaption,lipsum}
\usepackage{xcolor}
\usepackage{soul}                                     
\usepackage{caption}
\usepackage{algorithmic}
\usepackage{array}
\usepackage{textcomp}
\usepackage{stfloats}
\usepackage{url}
\usepackage{verbatim}
\usepackage{amsmath,amsfonts}
\usepackage{balance}
\usepackage{booktabs}
\usepackage{tikz}
\usetikzlibrary{arrows.meta, positioning, shapes, calc}
\usepackage{colortbl}
\usepackage{xcolor}
\usepackage{makecell} % Add this in the preamble if not already
\usepackage{enumitem}
\usepackage{multirow}
\title{\LARGE \bf 
RoboSubtaskNet: Temporal Sub-task Segmentation for Human-to-Robot Skill Transfer in Real-World Environments

\author{ Dharmendra Sharma, Archit Sharma, John Rebeiro, Vaibhav Kesharwani, Peeyush Thakur, \\Narendra Kumar Dhar, and Laxmidhar Behera\\
}}

\begin{document}
\maketitle

\begin{abstract}
Temporally locating and classifying fine-grained sub-task segments in long, untrimmed videos is crucial to safe human–robot collaboration. Unlike generic activity recognition, collaborative manipulation requires sub-task labels that are directly \emph{robot-executable}. We present RoboSubtaskNet, a multi-stage human-to-robot sub-task segmentation framework that couples attention-enhanced I3D features (RGB+optical flow) with a modified MS-TCN employing a Fibonacci dilation schedule to capture better short-horizon transitions such as reach–pick–place. The network is trained with a composite objective comprising cross-entropy and temporal regularizers (truncated MSE and a transition-aware term) to reduce over-segmentation and to encourage valid sub-task progressions. To close the gap between vision benchmarks and control, we introduce RoboSubtask, a dataset of healthcare and industrial demonstrations annotated at the sub-task level and designed for deterministic mapping to manipulator primitives. Empirically, RoboSubtaskNet outperforms MS-TCN and MS-TCN++ on GTEA and our RoboSubtask benchmark (boundary-sensitive and sequence metrics), while remaining competitive on the long-horizon Breakfast benchmark. Specifically, RoboSubtaskNet attains F1@50 = $79.5\%$, Edit = $88.6\%$, Acc = $78.9\%$ on GTEA; F1@50 = $30.4\%$, Edit = $52.0\%$, Acc = $53.5\%$ on Breakfast; and F1@50 = $94.2\%$, Edit = $95.6\%$, Acc = $92.2\%$ on RoboSubtask. We further validate the full perception-to-execution pipeline on a 7-DoF Kinova Gen3 manipulator, achieving reliable end-to-end behavior in physical trials (overall task success approx $91.25\%$). These results demonstrate a practical path from sub-task level video understanding to deployed robotic manipulation in real-world settings.
\end{abstract}


\pagenumbering{gobble}
%%%%%%%%%%%%%%%%%%%%%%%%%%%%%%%%%%%%%%%%%%%%%%%%%%%%%%%%%%%%%%%%%%%%%%%%%%%%%
\section{Introduction}
Deployment of robots in real-world scenarios, particularly in healthcare \cite{healthcare} and industrial \cite{industrial} settings, is no longer limited to repetitive or isolated automation tasks. Instead, robots are increasingly expected to operate collaboratively alongside humans, facilitating complex workflows that include interactive tasks such as handovers, pick and place operations, pouring liquids into containers, and cleaning work surfaces. These collaborative settings demand precision, reliability, and adaptability to human actions and environmental variability, making the design of robust perception and execution frameworks a critical requirement. Unlike full human activity recognition (HAR), these domains require decomposition of complex demonstrations into \textit{fine-grained robot-executable sub-tasks}. Bridging human video demonstrations with reliable robotic execution requires (i) \textit{robust spatiotemporal perception}, (ii) \textit{precise temporal segmentation into primitives}, and (iii) a \textit{deterministic mapping from predicted sub-task labels to robot skills}.  

Research on human-robot collaboration has led to different approaches that improve interaction schemes and execution efficiency. Han \textit{et al.}~\cite{Han_ICRA} and Hawkins \textit{et al.}~\cite{hawkins_2014} employed probabilistic graphical models to estimate task progress by jointly tracking hand and object positions. Brooks \textit{et al.}~\cite{IROS_2018} developed proactive robot assistants with multimodal recognition, while Fishman \textit{et al.}~\cite{ICRA_2020} introduced collaborative interaction models that improved the efficiency of human–robot learning. Johns \textit{et al.}~\cite{Johns_2021} proposed coarse-to-fine imitation learning, enabling robots to perform manipulation tasks from a single demonstration. Similarly, Semeraro \textit{et al.}~\cite{Francesco_2023} surveyed the integration of machine learning in collaborative robotics, and Thumm \textit{et al.}~\cite{ICRA_2024} introduced the “Human-Robot Gym” to benchmark reinforcement learning algorithms from demonstrations.  
\begin{figure}[H]
    \centering
    \fbox{\includegraphics[width=0.98\linewidth]{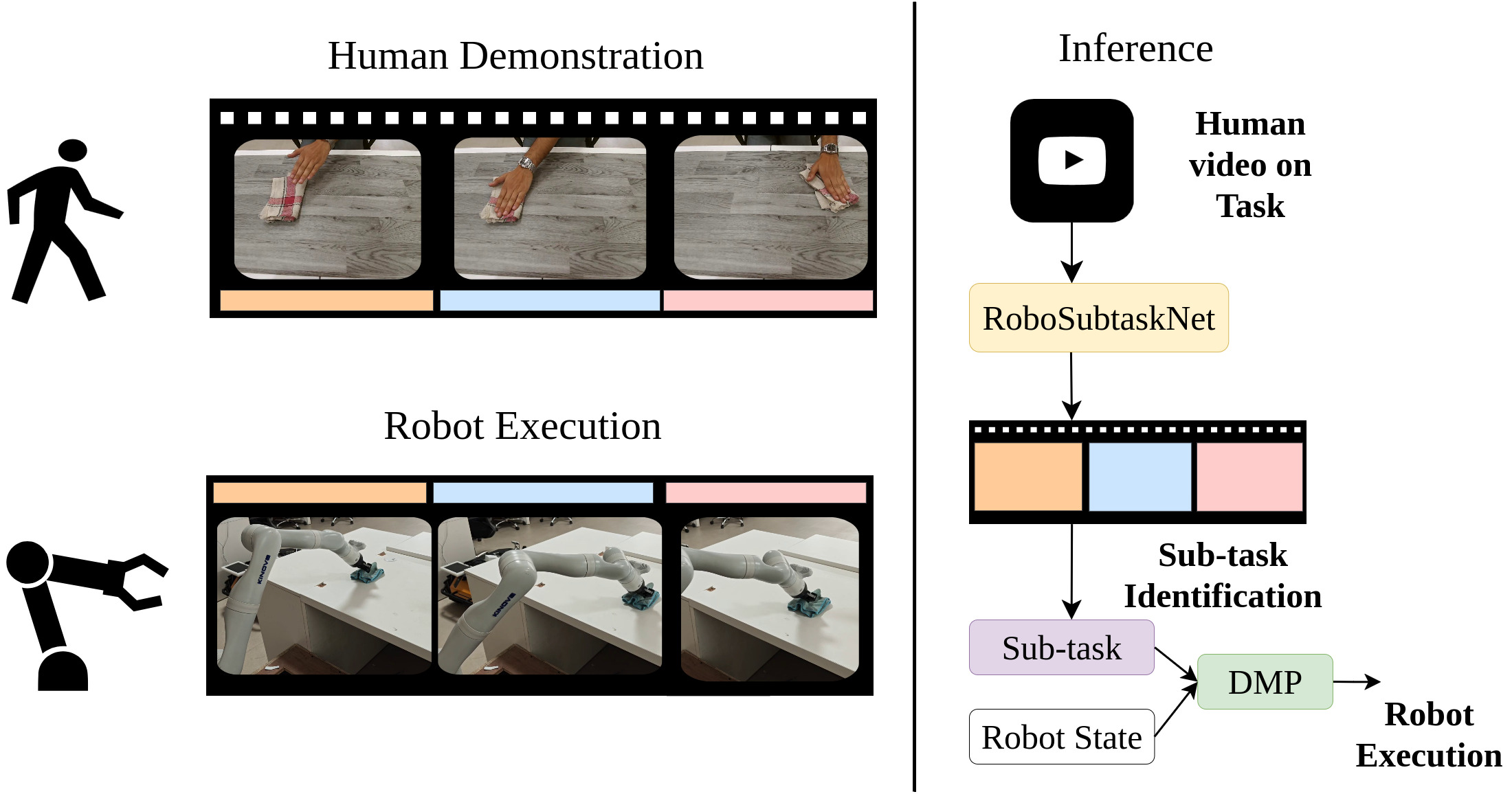}}
    \caption{RoboSubtaskNet pipeline for Human-to-Robot skill transfer.}
    \label{fig:workflow_diagram}
\end{figure}

Despite these advances, segmenting full task human videos into \emph{robot-executable sub-tasks} remains a challenge. Vision-based HAR datasets such as GTEA or EGTEA~\cite{fathi_eccv12_gtea}, Breakfast~\cite{kuehne_cvpr14_breakfast}, and 50Salads~\cite{stein_cviu16_50salads} provide valuable benchmarks for activity parsing but are not aligned with robotic execution needs. Their annotations focus on high-level semantic activities (e.g., “make tea”), while manipulator control requires primitive-level instructions such as \emph{reach}, \emph{pick}, or \emph{place}. As existing HAR datasets are mismatched for manipulator-level execution, we create \textit{RoboSubtask}, which contains
%healthcare and industrial 
human demonstrations 
%annotated 
with sub-task level labels explicitly designed for robot mapping. To ensure alignment with robotic control, our dataset defines sub-tasks that are directly mappable to manipulator primitives, as summarized in Table~\ref{tab:subtask}.  
\begin{table}[h!]
\caption{Definition of some fundamental sub-tasks}
  \centering
\begin{tabular}{ |p{1.5 cm}|p{6 cm}|}
\hline
\textbf{Sub-task}  & \textbf{Definition} \\
\hline
Reach & Moving the hand or end-effector towards an object \\
Pick  & Lifting an object at a certain height \\
Place & Positioning a picked object at a target location \\
Retract & Returning to home position \\
Pour & Pouring contents from one object into another \\
Wipe  & Cleaning the work surface \\
Move & Movement of the hand with objects\\
\hline
\end{tabular}
  \label{tab:subtask}
\end{table} 
Temporal action segmentation methods have made notable progress in parsing long and untrimmed videos. MS-TCN~\cite{farha_cvpr19_mstcn} and MS-TCN++~\cite{li_tpami20_mstcnpp} introduced dilated temporal convolutions and stage-wise refinement to improve segmentation quality, while Yi \textit{et al.}~\cite{yi_bmvc21_asformer} and Ding \textit{et al.}~\cite{ding_cvpr18_isba} further enhanced temporal modeling using self-attention and boundary-aware mechanisms. However, these models are primarily evaluated on video classification benchmarks rather than robotic execution pipelines. They are also more computationally expensive than TCNs, and their dilation schedules are optimized for long-horizon activities such as cooking, rather than the short-horizon sub-tasks required in manipulation. 

To overcome these gaps, we propose a novel framework, \textit{human-to-robot sub-task segmentation}, that integrates I3D-based spatiotemporal feature extraction~\cite{carreira_cvpr17_i3d} with a modified MS-TCN for fine-grained segmentation on manipulators.

The key \textbf{contributions} of this work are as follows:  
\begin{itemize}
    \item A feature-enhanced I3D backbone is designed by appending an attention fusion layer, enabling more discriminative spatio-temporal representations and reliable recognition of subtle human motions.  
    \item A fibonacci-dilated MS-TCN is proposed, which effectively models short-horizon transitions (e.g., reach-pick-place) in manipulation tasks compared to conventional dilation schedules.  
    \item A new dataset, RoboSubtask, is introduced, comprising healthcare and industrial demonstrations annotated at the sub-task level and directly mappable to manipulator primitives.  
    \end{itemize}
 The framework is validated through experiments on a Kinova Gen3 manipulator, demonstrating end-to-end execution from human video input to sequential primitive control.  
  
The workflow of the proposed framework is shown in Figure \ref{fig:workflow_diagram}. Our focus is on short-horizon manipulation sub-tasks in healthcare and industrial settings; extension to more complex long-horizon activities is left for future work.

The rest of the paper is organized as follows: Section II presents the proposed methodology, Section III discusses the experimental results, and Section IV concludes the paper.  
%%%%%%%%%%%%%%%%%%%%%%%%%%%%%%%%%%%%%%%%%%%%%%%%%%%%%%%%%

\section{Proposed Methodology}
The proposed methodology comprises four phases: (a) \textit{dataset creation}, (b) \textit{feature extraction and modality fusion}, (c) \textit{sub-task learning}, and (d) \textit{robot execution model}.

% \subsection{Dataset creation}\label{sec:pre-process}
% Human demonstration videos are recorded for each task using the Intel RealSense Depth Module (D410), and there are four tasks \textit{Pick and Place}, \textit{Pick and Pour}, \textit{Pick and Give}, and \textit{table Cleaning} in our RoboSubtask dataset. For each task, we acquire $200$ labeled videos. We hold out $20$\% ($40$ videos) per task as an unaltered validation/test set and use the remaining $160$ videos per task for augmentation. Two augmentations are applied independently: the first one is a horizontal flip to mitigate handedness/side bias and improve viewpoint invariance, and the second one is brightness adjustment to increase robustness to illumination changes (shadows, glare, under/over-exposure). This yields ($160 \times (1+2)=480$) training videos per task, totaling ($4 \times 480 = 1{,}920$) training videos across all tasks. The held-out set contains $40$ videos per task ($160$ total) for evaluation on unseen data.
% The supervised temporal annotation of every frame for each video is done manually for segmenting the subtask in each video. The sequence of subtasks associated with each task is shown in Table \ref{tab:subtasks_list}. Annotations are stored per video as temporal label sequences to enable flexible downstream processing. The data set spans multiple users, objects, and environments to promote generalization.

%%%%%%%%%%%%%%%%In_general_terms%%%%%
\subsection{Dataset creation}\label{sec:pre-process}
We consider a dataset of $T$ manipulation tasks $\mathcal{T}=\{t_1,\ldots,t_T\}$ recorded with an RGB  vision sensor mounted on the robot. For each task $t\in\mathcal{T}$, we acquire $N_t$ labeled videos. Each video is temporally annotated at the \textit{frame} level manually with labels of a sub-task from vocabulary $V$ (a task to sub-task sequence summarized in Table~\ref{tab:subtasks_list}); annotations are stored per video as a temporal label sequence for flexible downstream processing. We create a hold-out split by reserving a fraction $r_{\mathrm{val}}$ of videos per task for validation and testing, leaving $(1-r_{\mathrm{val}})N_t$ for training. Let ${A}=\{a_1,\ldots,a_{|{A}|}\}$ denote a set of augmentations that preserve appearance and geometry applied independently to each training video; the effective multiplier of augmentations is $(1+|{A}|)$. The per-task and total training sizes are therefore
$N^{\mathrm{train,aug}}_{t} \;=\; (1-r_{\mathrm{val}})\,N_t\,\bigl(1+|{A}|\bigr), N^{\mathrm{train,aug}} \;=\; \sum_{t\in\mathcal{T}} N^{\mathrm{train,aug}}_{t},$ and the holdout size per task is $N^{\mathrm{val}}_{t}=r_{\mathrm{val}}N_t$. Data collection spans multiple users, objects, and environments to promote generalization before feature extraction.

\begin{table}[!ht]
\centering
\caption{Task to sub-task mapping used for annotation and control on RoboSubtask (Columns indicate sub-task vocabulary. \checkmark denotes sub-tasks in each task’s sequence.)}

\label{tab:subtasks_list}
\resizebox{\columnwidth}{!}{%
\begin{tabular}{lcccccccccc}
\toprule
\textbf{Task} & 
\textbf{Reach} & 
\textbf{Pick} & 
\textbf{Move} & 
\textbf{Pour} &
\textbf{Move} &
\textbf{Give} &
\textbf{Place} & 
\textbf{Wipe} & 
\textbf{Retract} \\
\midrule
Pick \& Place & 
\checkmark & \checkmark & \checkmark &  &     &  &   \checkmark & & \checkmark\\

Pick \& Pour & 
\checkmark & \checkmark & \checkmark & \checkmark &  \checkmark &  & \checkmark& &  \checkmark\\

Cleaning & \checkmark& & & & & & & \checkmark &\checkmark\\

Pick \& Give & 
\checkmark & \checkmark &  &  &    & \checkmark &   &  &\checkmark\\

\bottomrule
\end{tabular}%
}
\end{table}
%%%%%%%%%%%%%%%%%%%%%%%%%%%%%%%%%%%%%%%%%%%%%%%%%%%%%%%%%%%%%%%%%%%
\subsection{Feature extraction and modality fusion}
\label{subsec:feature_extraction}
\paragraph{I3D for video features}
Action segmentation requires representations that capture appearance and motion. Inflated 3D ConvNets (I3D) \cite{I3D} learn spatiotemporal features directly from short clips, going beyond framewise 2D CNNs by modeling temporal evolution. Pre-training on a large video kinetics dataset \cite{kinetics_dataset} makes I3D an effective standard feature encoder for human demonstrations with limited task-specific labels. In addition, downstream temporal models such as TCN and MSTCN operate on compact, high-level descriptors rather than raw pixels; I3D supplies precisely these per-time-step embeddings, improving segmentation robustness and data efficiency.

\begin{figure*}[!htb]
  \centering
  \begin{subfigure}[b]{0.45\linewidth}
    \centering
    \includegraphics[width=\linewidth]{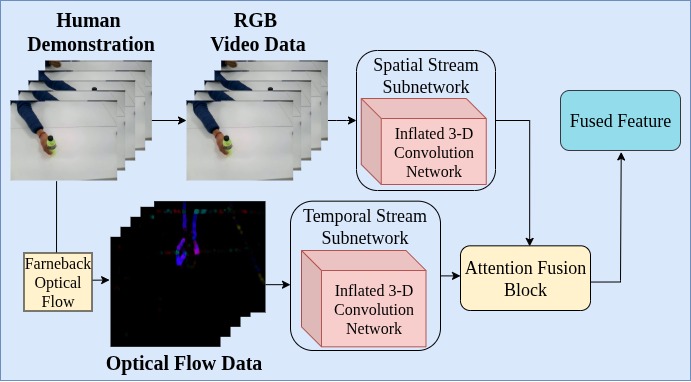}
    \caption{
    % Feature extraction using I3D with attention fusion layer.
    }
    \label{fig:I3D_base_1}
  \end{subfigure}
  \hfill
  \begin{subfigure}[b]{0.53\linewidth}
    \centering
    \includegraphics[width=\linewidth]{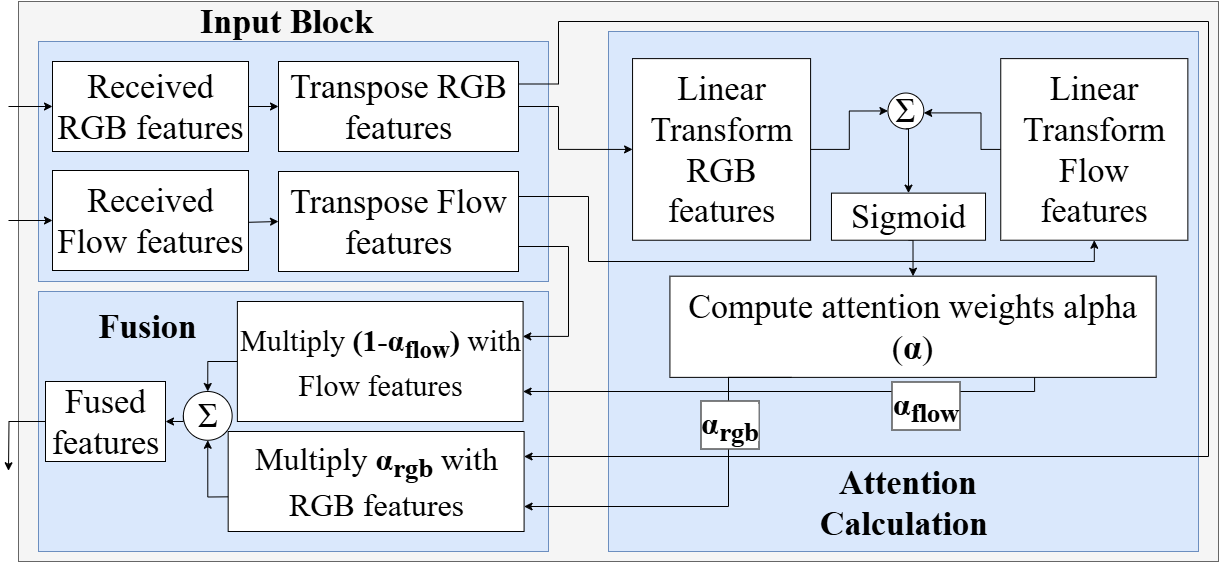}
    \caption{
    % A schematic of attention fusion.
    }
    \label{fig:I3D_attention_fusion}
  \end{subfigure}

  \caption{Feature extraction and attention-fusion module. (a) I3D-based feature extractor, (b) Attention fusion schematic.}
  \label{fig:two_subfigures}
\end{figure*}

\paragraph{Raw video to feature matrices.}
Let a demonstration video contain $F$ RGB frames. We derive two frame sequences (shown in Figure \ref{fig:I3D_base_1})
\[
\{F^{\text{rgb}}_i\}_{i=1}^{F} \quad \text{and} \quad \{F^{\text{flow}}_i\}_{i=1}^{F},
\]
where $F^{\text{flow}}_i$ denotes the optical-flow representation between frames $i$ and $i{+}1$. Passing each sequence through I3D heads yields \textit{D} (say $1024$) descriptors per time step as
\begin{align*}
\mathbf{f}^{\text{rgb}}_t = \phi^{\text{rgb}}_{\text{I3D}}\!\big(F^{\text{rgb}}_t\big) \in \mathbb{R}^{D}\quad\text{and}\quad 
\mathbf{f}^{\text{flow}}_t = \phi^{\text{flow}}_{\text{I3D}}\!\big(F^{\text{flow}}_t\big) \in \mathbb{R}^{D},
\end{align*}
for $t=1,\dots,T$. I3D includes temporal down-sampling (strided 3D convolutions and pooling) and $T \approx \lceil F/s \rceil$ with temporal stride $s\!\approx\!8$. For efficient and archival reuse, we store
\[
\mathbf{X}^{\text{rgb}} = \big[\mathbf{f}^{\text{rgb}}_1,\dots,\mathbf{f}^{\text{rgb}}_T\big]^\top \in \mathbb{R}^{T \times D} \,\,\,\text{and}\]
\[\mathbf{X}^{\text{flow}} = \big[\mathbf{f}^{\text{flow}}_1,\dots,\mathbf{f}^{\text{flow}}_T\big]^\top \in \mathbb{R}^{T \times D}
\]
for downstream training.

% \begin{figure}
%     \centering
%     \includegraphics[width=1\linewidth]{I3D-modified.drawio-1.pdf}
%     \caption{Feature extraction using I3D with an extra attention fusion layer.}
%     \label{fig:I3D_base_1}
% \end{figure}
% %\vspace{0.1em}
% \begin{figure}
%     \centering
% \includegraphics[width=1\linewidth]{fusion attension.pdf}
%     \caption{A schematic of attention fusion.}
%     \label{fig:I3D_attention_fusion}
% \end{figure}

\paragraph{Attention-based fusion of RGB and Flow features}
Once we have extracted the RGB features $f^{rgb}_t$ and flow features $f^{flow}_t$ from the I3D model, the next challenge is to combine them effectively. A straightforward approach would be to concatenate them into a single $D+D$-dimensional feature vector per frame. However, such fixed fusion treats both modalities as equally important at all times, which is often not true in practice.

For instance, when the hand is in motion, optical flow may carry more meaningful cues, whereas during precise object interactions (like grasping), RGB appearance becomes more critical. To address this imbalance, we introduce an attention-based fusion mechanism shown in Figure \ref{fig:I3D_attention_fusion} that allows the model to adaptively assign importance to each modality for every frame and every dimension of the feature. It uses a small neural network that takes RGB and flow as input and predicts attention weights for each frame.
Formally, for each frame $t$, a fully connected shallow layer computes an attention coefficient $\alpha(t) \in [0,1]^{D}$ as
\begin{align}
\alpha(t) &= \sigma\big(W_a \cdot [f^{rgb}_t ; f^{flow}_t] + b_a\big),
\end{align}
where $W_a$ and $b_a$ are learnable parameters and $\sigma(\cdot)$ denotes sigmoid activation. The fused feature is computed as
\begin{align}
f^{fused}_t &= \alpha(t) \odot f^{rgb}_t + \big(1 - \alpha(t)\big) \odot f^{flow}_t.
\end{align}
Here, $\odot$ denotes element-wise multiplication. Intuitively,
if $\alpha(t) \approx 1$, the model is based on RGB whereas for $\alpha(t) \approx 0$, the model focuses on flow. In most cases, $\alpha(t)$ learns nuanced combinations, capturing both appearance and motion cues.

The network design allows for dynamic emphasis on the most informative modality depending on the sub-task. For example, \textit{reach} and \textit{move} are dominated by motion cues (flow), while \textit{pick} or \textit{place} often require visual detail (RGB). Table~\ref{tab:subtask-dominance} illustrates typical dominant features in different sub-tasks.

\begin{table}[h]
\caption{Dominant feature modality across sub-tasks}
  \centering
\begin{tabular}{ |p{1.5 cm}|p{6 cm}|}
\hline
\textbf{Sub-task}  & \textbf{Likely dominant feature} \\
\hline
Reach & Flow (motion of hand) \\
Pick  & RGB (appearance of fingers around object) \\
Place & RGB (object shape on surface) \\
Retract & Flow (hand moving away) \\
Pour & RGB \\
Wipe  & Flow \\
Move  & Flow \\
\hline
\end{tabular}
  \label{tab:subtask-dominance}
\end{table} 

Compared to concatenation or simple averaging, attention fusion provides several advantages. With adaptive weighting, it learns the relative importance of RGB and flow. The weighting changes across time and sub-tasks, making the representation more robust. The dimension of the fused representation is $D$ instead of $2D$, thus reducing computational cost. By focusing on the right modality at the right time, the model achieves better temporal consistency and accuracy in sub-task recognition. Therefore, attention-based fusion allows the system to intelligently decide “which modality matters more” at each step of the task, resulting in stronger and more interpretable feature embeddings for downstream segmentation. Following feature extraction from the I3D model with an attention fusion block, the fused features are fed into the sub-task segmentation network. In contrast to the exponential dilation schedule used in MS-TCN (i.e., $d_l=2^{l-1}$), we employ a Fibonacci dilation scheme that is better suited to short-horizon manipulation transitions.
%%%%%%%%%%%%%%%%%%%%%%%%%%%%%%%%%%%%%%%%%%%%%%%%%%%%%

\subsection{Sub-task learning}\label{subtask_learning}
Understanding complex human activities requires modeling long-range temporal structure beyond the per-frame features. While attention-enhanced I3D (RGB+Flow) provides rich local spatiotemporal descriptors, human demonstrations are inherently sequential, consisting of interdependent sub-tasks (e.g., \textit{reach, pick, pour, place, retract}). We, therefore, adopt a multistage temporal convolutional architecture that captures long-range dependencies and progressively refines boundaries. Our contributions are twofold: (i) a \textit{Fibonacci-based dilation schedule} and (ii) a \textit{transition-aware loss} that discourages invalid label switches. 

%The notations used in this section
Here, $\mathbf{X}\in\mathbb{R}^{T\times D}$ denotes the fused feature sequence
%(Sec.~\ref{subsec:feature_extraction}), 
with $T$ time steps and $D$ channels. A stage applies $L$ dilated layers with kernel size $k$ and dilation factors \(\{d_l\}_{l=1}^L\). We denote the hidden sequence after layer \(l\) as \({H}^{(l)}\in\mathbb{R}^{T\times D}\) (with \(\mathbf{H}^{(0)}{=}\mathbf{X}\)).
\par
\subsubsection{Single stage TCN}
Each layer is a dilated $1$D temporal convolution with a residual connection:
\begin{equation}
{H}^{(l)}_t \;=\; \mathrm{ReLU}\!\big(\big({W}^{(l)} *_{d_l} {H}^{(l-1)}_t\big) + {b}^{(l)}\big) \;+\; {H}^{(l-1)}_t,
\end{equation}
where \( *_{d_l} \) denotes convolution along time with dilation \(d_l\), \({W}^{(l)}\in\mathbb{R}^{D\times D\times k}\) are the layer weights, and \({b}^{(l)}\in\mathbb{R}^{D}\) is a bias. With \(k{=}3\), each layer expands the receptive field (RF) by \(k{-}1{=}2\) time steps.

\begin{figure}[H]
    \centering
    \includegraphics[width=\columnwidth]{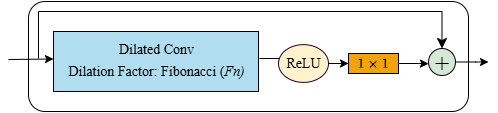}
    \caption{Dilated residual layer with Fibonacci dilation factors.}
    \label{fig:modified_TCN_residual}
\end{figure}

\paragraph{Fibonacci dilated convolutions}
Conventional temporal segmentation models employ exponentially increasing dilation factors such as $d_l{=}2^{\,l-1}$ (e.g., $1,2,4,8,\ldots$). It also creates a sparse sampling pattern, skipping over important intermediate frames and causing the model to miss subtle transitions. To overcome this, we introduce a Fibonacci dilation strategy.
\begin{equation}
d_l \;=\; F_{l+1}, \qquad l=1,\dots,L,
\end{equation}
where \(F_{l+1}\) are Fibonacci numbers \((1,1,2,3,5,\ldots)\). With $k$ kernel as $3$, the receptive field after $L$ layers is
\begin{equation}
\mathrm{RF} \;=\; 1 + (k-1)\sum_{l=1}^{L}F_{l+1} \;=\; 1 + 2\big(F_{L+3}-2\big),
\end{equation}
yielding denser temporal coverage of short-horizon events while still expanding context for move-like segments. An illustrative residual layer is shown in Figure~\ref{fig:modified_TCN_residual}.

\subsubsection{Frame wise classification}
A \(1{\times}1\) temporal convolution maps \({H}^{(L)}_t\) to class logits, and a softmax yields per-frame probabilities:
\begin{equation}
{P}_t \;=\; \mathrm{Softmax}\!\big({W}_o\,{H}^{(L)}_t + {b}_o\big), \qquad
{W}_o\in\mathbb{R}^{C\times D},\ {b}_o\in\mathbb{R}^{C},
\end{equation}
where $C$ is the number of sub-task classes and \({P}_t\in\mathbb{R}^C\) is on the probability simplex.

\subsubsection{Multi-stage refinement}
We stack \(S\) stages; stage \(s\) refines the predictions of stage \(s{-}1\),
\begin{align}
{Y}^{(1)} &= F({X}),\\
{Y}^{(s)} &= F\!\big({Y}^{(s-1)}\big), \qquad s=2,\dots,S,
\end{align}
where \(F(\cdot)\) denotes a single-stage TCN (weights are not shared between stages). This progressively reduces over-segmentation and sharpens boundaries.

\subsubsection{Loss Function}
We minimize the combined loss function
\begin{equation}
\mathcal{L} \;=\; \sum_{s=1}^{S} \Big({L}_{\mathrm{CE}}^{(s)} + \lambda\, {L}_{\mathrm{T\text{-}MSE}}^{(s)} + \gamma\, {L}_{\mathrm{Trans}}^{(s)}\Big),
\end{equation}
with nonnegative weights \(\lambda,\gamma\) and truncation \(\tau>0\). Some loss functions used in our framework are:

\paragraph{Cross entropy}
For ground-truth labels \(c_t\in\{1,\dots,C\}\),
\begin{equation}
{L}_{\mathrm{CE}}^{(s)} \;=\; -\frac{1}{T}\sum_{t=1}^{T} \log {P}_t[c_t].
\end{equation}

\paragraph{Temporal smoothing (T-MSE)}
To reduce flicker and over-segmentation, we penalize abrupt changes in log-probabilities:
\begin{equation}
{L}_{\mathrm{T\text{-}MSE}}^{(s)} \;=\; \frac{1}{T-1}\sum_{t=2}^{T}\min\!\Big(\big\|\log {P}_t - \log {P}_{t-1}\big\|_2^2,\, \tau\Big).
\end{equation}

\paragraph{Transition aware}
Let \(M\in\{0,1\}^{C\times C}\) encode \emph{invalid} transitions (\(M_{ij}{=}1\) if \(i{\rightarrow}j\) is invalid, else \(0\)). Define a transition weight
\begin{equation}
w_t \;=\; {P}_{t-1}^{\top}\, M\, {P}_t \;\in [0,1],
\end{equation}
 modulate the temporal change accordingly;
\begin{equation}
{L}_{\mathrm{Trans}}^{(s)} \;=\; \frac{1}{T-1}\sum_{t=2}^{T} w_t \,\big\|{P}_t - {P}_{t-1}\big\|_1.
\end{equation}
This discourages rapid switches that imply implausible sub-task progressions while leaving valid transitions less penalized.

% \subsubsection{Training  and implementation details}
% We use \(S{=}4\) stages; each stage has \(L{=}10\) dilated layers with \(k{=}3\), \(D{=}64\) channels, ReLU activations, and residual connections. We apply dropout \(p{=}0.5\) after temporal convolutions and \(\ell_2\) weight decay \(1{\times}10^{-4}\). Features are z-score normalized per channel over the training set. Class imbalance is addressed via inverse-frequency weights in \({L}_{\mathrm{CE}}\). We set \(\lambda{=}0.15\), \(\gamma{=}0.25\), and \(\tau{=}4.0\). Models are trained in PyTorch with AdamW (initial learning rate \(1{\times}10^{-3}\)), cosine decay with 5-epoch warmup, batch size 8 sequences, gradient clipping at 5.0, and mixed precision. Training runs for up to 60 epochs with early stopping on the validation Edit score; the checkpoint with the best F1@50 is reported. At inference, we apply repetition collapsing and an optional median filter (window 3) to smooth isolated spikes.

%%%%%%%%%%%%%%%%%%%%%%%%%%%%%%%%%%%%%%%%%%%%%%%%%%%%

\subsection{Robot execution model}
The robot execution model key components are discussed below.
\subsubsection{Learning trajectory using DMP}
Each sub-task (pick, retract, tilt, etc.) is predefined and trained using a DMP model. The DMP framework allows the creation of a library for these sub-tasks, facilitating smooth and human-like robot manipulator movements. DMP can be learned through demonstration or programmed manually to generate the appropriate trajectories. The DMPs in this work are formulated to generate trajectories in the Cartesian space for the robot's end-effector pose, specifically controlling its $x$, $y$, and $z$ coordinates.
To describe trajectory evolution, the DMP equation is represented as 
\begin{equation}
    \tau \ddot x = \alpha(\beta(g-x)-\tau \dot x)+ f(x),
    \label{eq:1}
\end{equation}
% $\tau \ddot x = \alpha(\beta(g-x)-\tau \dot x)+ f(x), $ 
  % $ $  \label{eq:1}
where $\tau$ is the temporal scaling factor that adjusts the duration of the movement. $x$, $\dot{x}$, and $\ddot{x}$ represent position, velocity, and acceleration, respectively. $g$ is the target position, with $\alpha$ controlling the speed of convergence to the target. $\beta$ controls the spring-like behavior of the trajectory, shaping the movement, and $f(x)$ is the forcing function that further refines the trajectory. The forcing function is represented as
\[
f(s) = \frac{\sum_{i=1}^{N} w_i \psi_i(s)}{\sum_{i=1}^{N} \psi_i(s)},  \psi_i(s) = \exp \left( - \frac{(s - c_i)^2}{2 \sigma^2} \right), \tau_c s' = -\alpha_x s,
\]
where 
%$f(s)$ is the forcing function, 
$w_i$ are weights, $\psi_i(s)$ is a basis function, $\tau_c$ is the temporal scaling factor, $s$ is the phase variable, and $\alpha_x$ is a time constant.
The detailed DMP trajectory learning is taken from \cite{DMP}.

\subsubsection{Object detection}
The vision-based grasping process begins with object detection. The robot captures an RGB image of the workspace and processes it through a YOLOv8-n model, which has been trained on custom objects (\textit{bottle, cup, bowl, medicine syrup, mop, box}) targeted for manipulation. Upon a successful detection, a bounding box is drawn around the object. The center pixel coordinates $(u,v)$ of this bounding box are calculated. These 2D image coordinates are then combined with data from a depth sensor, which provides a corresponding per-pixel distance map. Indexing this depth map at the location $(u,v)$ yields the distance $d$ from the camera to the object.

\subsubsection{Visual Servoing with a Proportional Controller} 
A proportional (P) controller is used to drive the end-effector from its current position to the target position. The components of this approach are control law design and implementation system constraints.

\paragraph{Control objective}
The main goal is to control the robotic arm's position in Cartesian coordinates ($x, y, z$) by minimizing position errors through velocity adjustments. A P-controller determines the difference between the desired and current positions, generating proportional velocity commands for the arm.

\paragraph{Proportional gains}
To ensure proper sensitivity of the controller, proportional gains \(k_x\) and \(k_{pz}\) are considered. \(k_x\) is used to control the arm's forward and backward movements in $x$-axis, while \(k_{pz}\) regulates movements in $y$ and $z$ directions.

\paragraph{Constraints}
To prevent too fast movement, velocity constraints are applied on all axes. When position errors are within defined thresholds, velocities are set to zero, allowing the arm to stabilize without unnecessary movements.

\paragraph{Control laws}
% For the $x$-axis, which typically controls the depth or distance from the object, the control law is
% $ 
% v_x = k_x  (d - d_{ref}),
% $
% where \(v_x\) is the velocity on the $x$-axis, \(k_x\) is the proportional gain, \(d\) is the measured depth from the object, and \(d_{ref}\) is the reference depth, typically set to $0.5$ meters. The control law for the $y$ and $z$ axes are $ 
% v_y = k_{pz} e_y \quad \text{and} \quad
% v_z = k_{pz} e_z
% $ 
% where \(v_y\) and \(v_z\) are velocities, \(e_y\) and \(e_z\) are position errors in the $y$ and $z$ directions, respectively.
The P-controller generates velocity commands proportional to position errors. For the $x$-axis (depth): $v_x = k_x\,(d - d_{\text{ref}})$, where $d$ is the measured depth and $d_{\text{ref}} = 0.5\ \mathrm{m}$. For the $y$ and $z$ axes: $v_y = k_{pz}\,e_y$ and $v_z = k_{pz}\,e_z$, where $e_y$ and $e_z$ are the position errors in $y$ and $z$, respectively. Velocity commands are saturated to respect axis limits and are set to zero when errors lie within the tolerance band.

% \subsubsection{Data integration}
% Depth measurements are incorporated into the control of movement of the $x$-axis, while coordinate data are used to refine the system’s understanding of the workspace. This ensures accurate adjustments to the positioning of the robotic arm during the grasping process.

% \subsubsection{Action execution}
% Finally, the computed velocity commands are published to the robotic system. The system continuously monitors the velocities of all three axes $(x, y, z)$. 

% Once the velocities in all three directions are zero, the depth measurement from the last control iteration is used to adjust the position on the $x$-axis. This ensures that the robotic arm moves to the correct depth to align with the object to be grasped. The system then publishes this last depth value as the $x$-axis command to the robot, guiding it precisely to the desired object.

\subsubsection{Coordinate Transformation and Task Execution via DMP}
The proportional controller continues until the positional error is minimized and the velocities for all axes converge to zero, indicating the end-effector is aligned with the target. Once the P-controller has achieved alignment, the final 3D coordinate of the object is transformed into the robot base frame using the ROS TF2 library to find the static transform between the camera and base frames. This transformed coordinate is passed as the goal $g$ to the DMP corresponding to the desired sub-task. DMP then generates a smooth trajectory from the current end-effector position to the target for final movement.
% The proportional controller operates until the positional error is minimized and the computed velocities for all axes converge to zero, indicating the end-effector is aligned with the target object. The final 3D coordinates $(x, y, z)$ from this converged state are then passed as the target goal to the appropriate DMP. The DMP subsequently generates a smooth trajectory from the current end-effector position to this new target. This hybrid approach ensures the robustness of a reactive vision-based controller for initial alignment and the precision and natural motion of a learned model for the final execution of the task, such as grasping or placement.

%%%%%%%%%%%%%%%%%%%%%%%%%%%%%%%%%%%%%%%%%%%%%%%%%%%%%%%%%%%%%%%%%%%
\begin{figure*}[!t]
  \centering
  \subfloat[Pick and place]{%
    \includegraphics[width=0.5\textwidth]{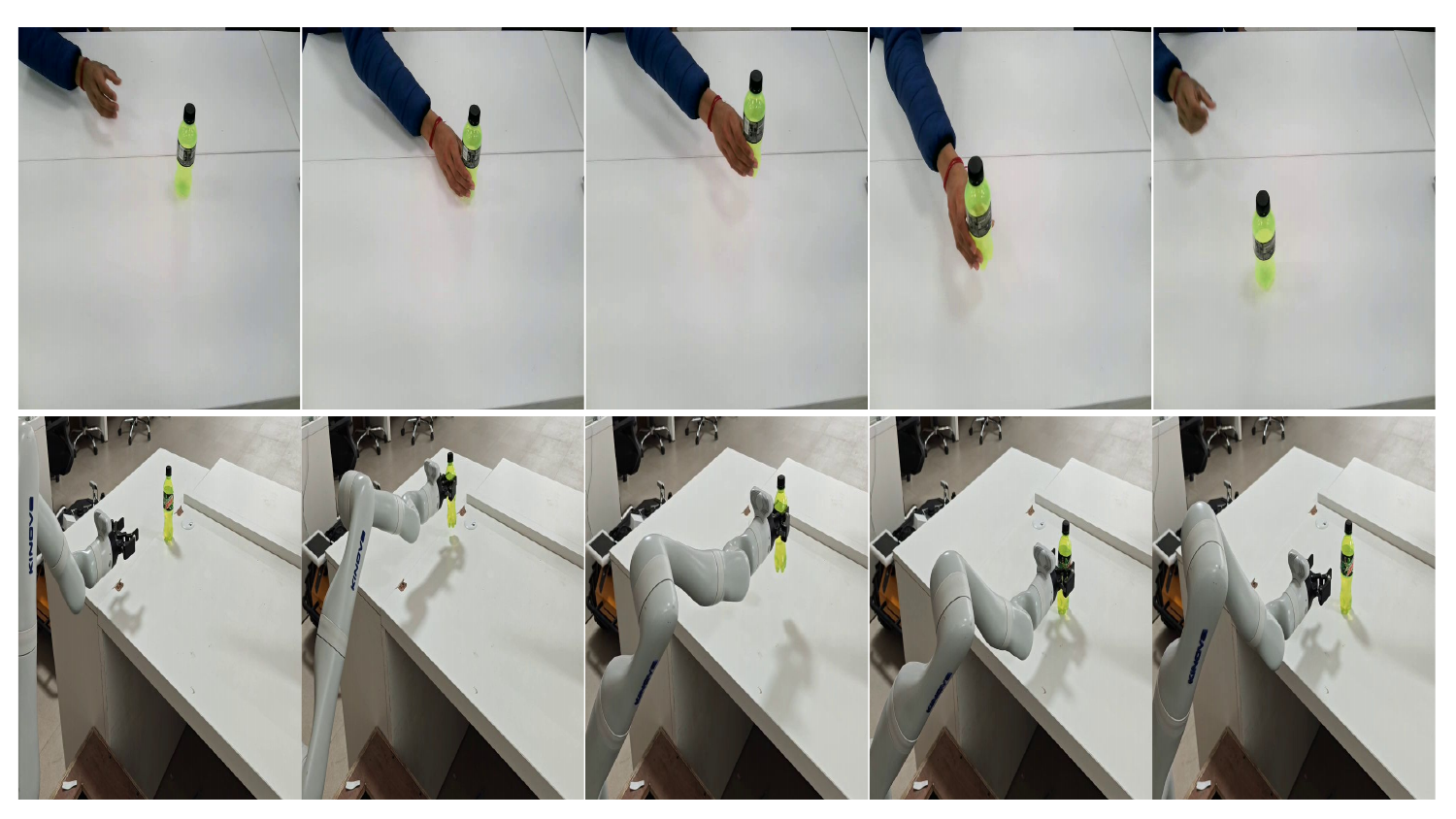}%
    \label{fig:pick_place}}
  \hfill
  \subfloat[Pick and pour]{%
    \includegraphics[width=0.5\textwidth,height=0.210\textheight]{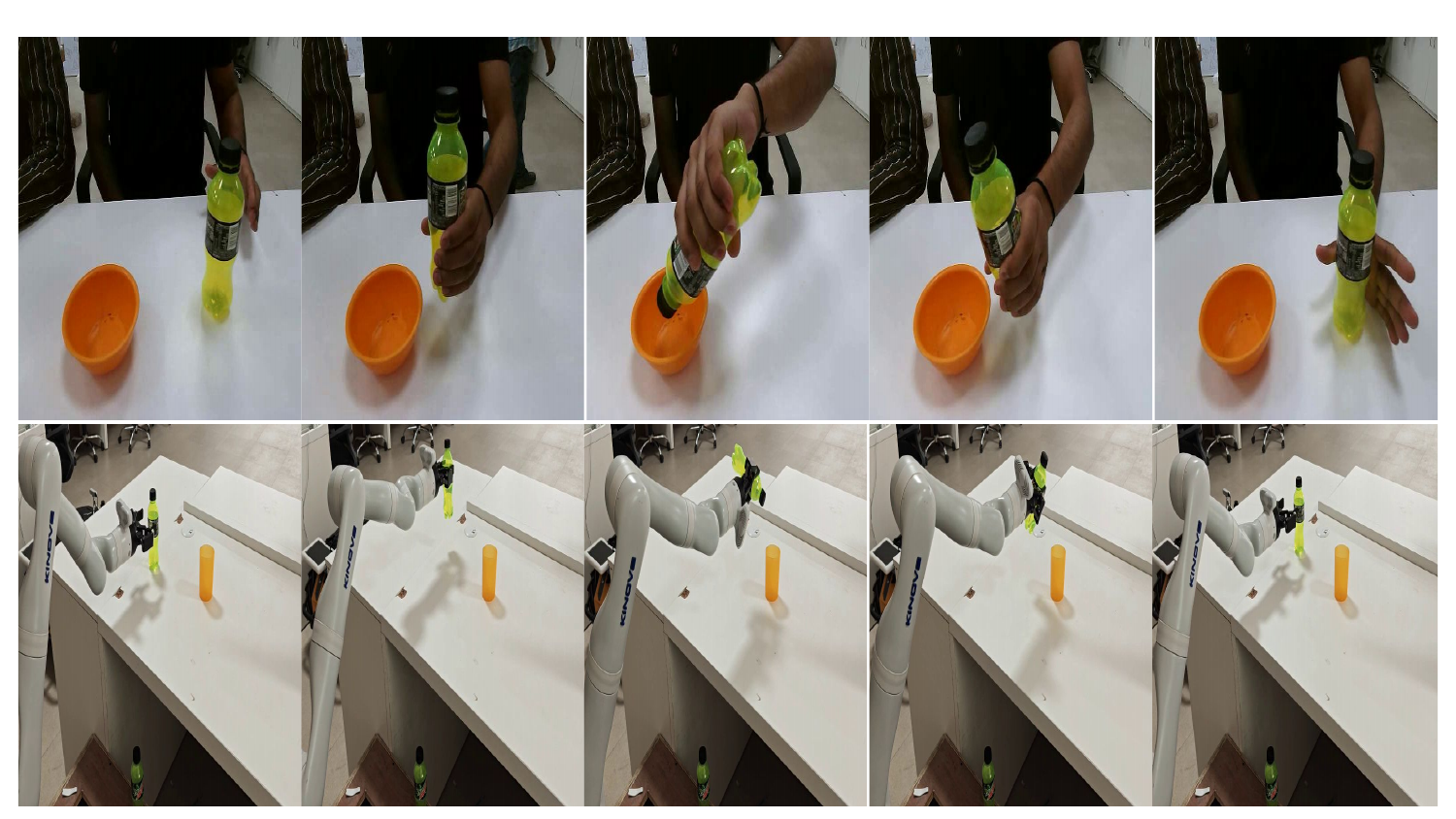}%
    \label{fig:pick_pour}}
  \vspace{0.1em}
  \subfloat[Cleaning of a table]{%
    \includegraphics[width=0.5\textwidth]{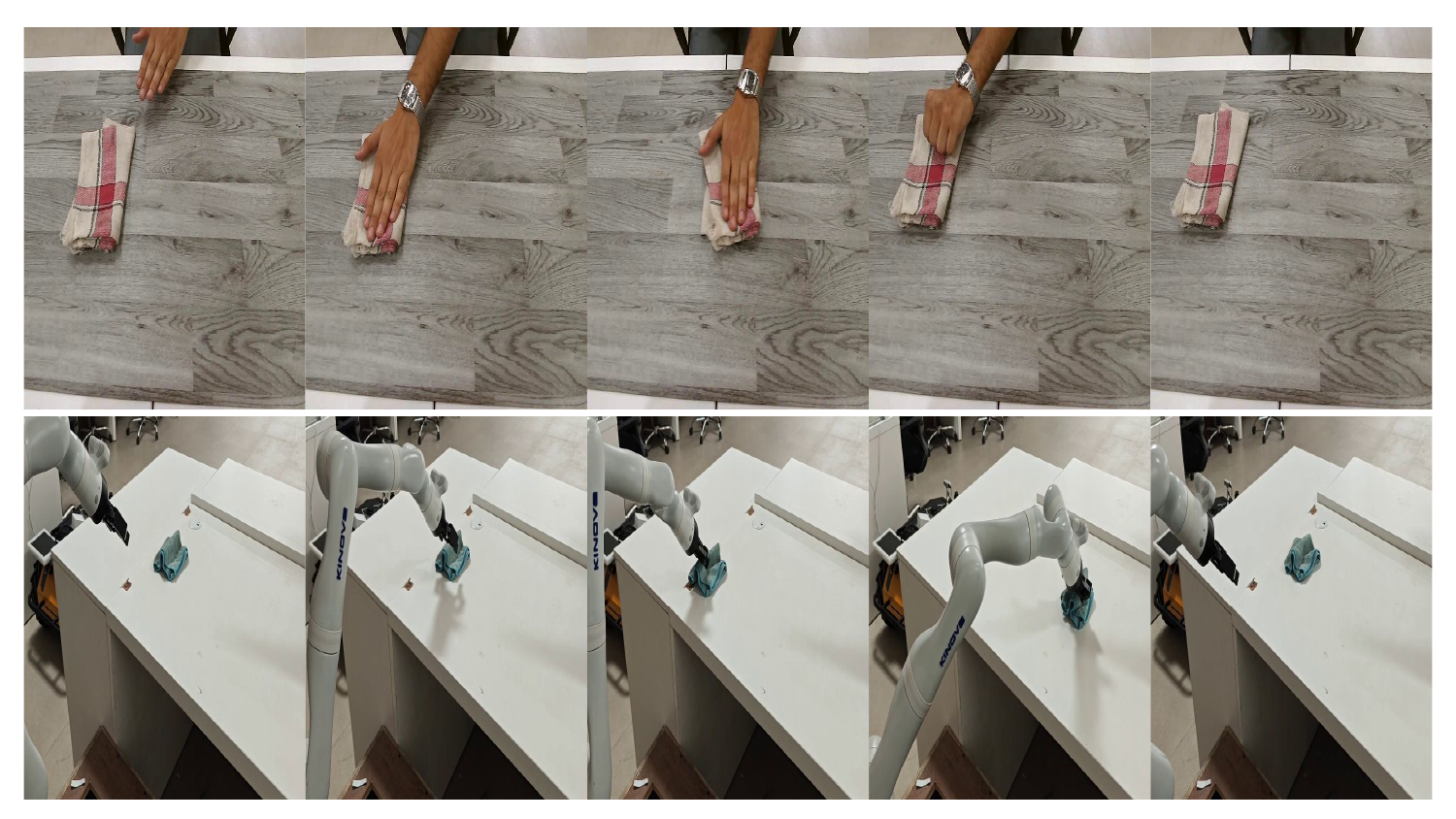}%
    \label{fig:clean_table}}
  \hfill
  \subfloat[Pick and give]{%
    \includegraphics[width=0.5\textwidth]{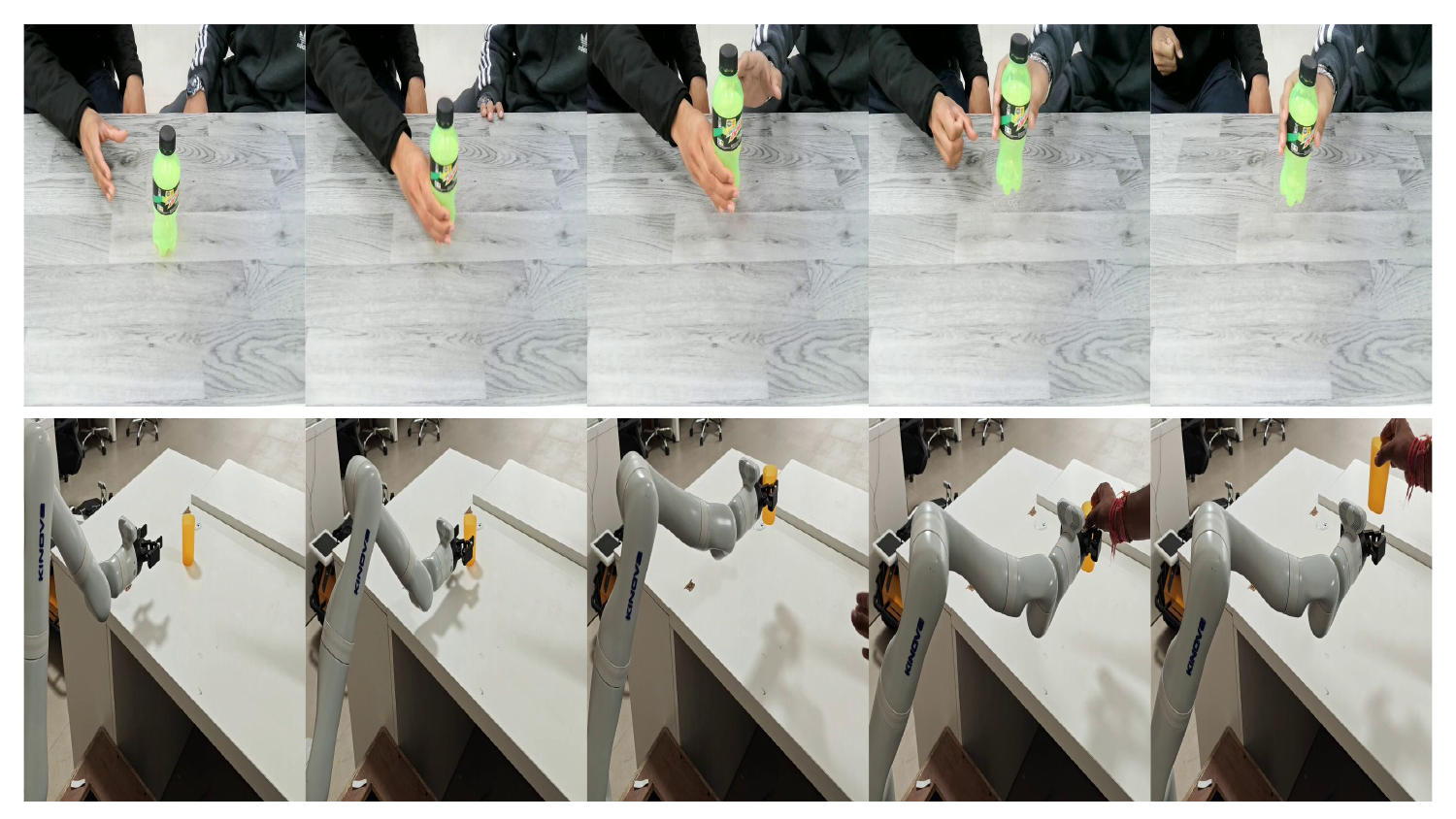}%
    \label{fig:pick_give}}

  \caption{For each task: (a) pick \& place, (b) pick \& pour, (c) table cleaning, \& (d) pick \& give. The top images show their human demonstrations, and the bottom images show the corresponding task sequence of the robot (sequence left to right).}
  \label{fig:task_grid}
\end{figure*}

%%%%%%%%%%%%%%%%%%%%%%%%%%%%%%%%%%%%%%%%%%%%%%%%%%%%%%%%%%%%%%%%%%%

\section{Experiments and Results}
We evaluated the proposed perception-to-execution pipeline on public temporal segmentation benchmarks and on our robotic platform with RoboSubtask, assessing both vision metrics and end-to-end task execution.

\subsection{Experimental setup}
The experimental setup consists of the following.
\subsubsection{Hardware} A Kinova Gen3 (7-DOF) robot manipulator executes the predicted sub-task sequence. An Intel RealSense depth camera (D410) is rigidly mounted on the arm for RGB-depth capture. Model training/inference is performed on a single NVIDIA GTX $1650$ ($4$GB VRAM).
\subsubsection{Software} We use robot operating system (ROS) with Kinova Kortex APIs~\cite{kinova_gen3_user_guide} for low-level control and vision acquisition. Temporal models are implemented in PyTorch.

\subsection{Datasets and metrics used}
\label{subsec:model_eval}
We evaluate the proposed framework on three challenging datasets: Georgia Tech Egocentric
Activities (GTEA) \cite{fathi_eccv12_gtea}, the Breakfast \cite{kuehne_cvpr14_breakfast}, and our RoboSubtask dataset.
\par
\subsubsection{RoboSubtask description} 
In our experiments, we use $T=4$ tasks, $\mathcal{T}$=(\textit{pick and place}, \textit{pick and pour}, \textit{pick and give}, \textit{cleaning}) recorded with RGB camera. For each task, $N_t=200$; the holdout fraction is $r_{\mathrm{val}}=0.20$ (unaltered). The augmentation set is ${A}$=(\textit{horizontal flip}, \textit{brightness adjustment}) so $|A|=2$ and the multiplier is $1+|A|=3$. Hence,
$N^{\mathrm{train,aug}}_{t}=(1-0.20)\times 200 \times 3 = 480\ \text{videos per task}, N^{\mathrm{train,aug}}=4\times 480=1{,}920,
$
and \(N^{\mathrm{val}}_{t}=0.20\times 200=40\), totaling \(160\) hold-out videos across tasks. All videos are frame-wise annotated using the sub-task vocabulary in Table~\ref{tab:subtasks_list}.

Following the MS-TCN metrics~\cite{farha_cvpr19_mstcn,li_tpami20_mstcnpp}, we use \emph{frame-wise accuracy}, segmental \emph{F1@}\{10, 25, 50\} (IoU thresholds), and \emph{Edit} score to evaluate the performance of various methods in the data sets mentioned above. Accuracy captures per-frame correctness; $F1$ emphasizes boundary quality; Edit penalizes over-segmentation and order errors.

%%%%%%%%%%%%%%%%%%%%%%%%%%%%%%%%%%%%%%%%%%%%%%%%%%%%%
\subsection{Training and implementation details}
We train a multi-stage temporal model with $S$ stages; each stage comprises $L$ 1D convolutional layers with a dilation schedule $d_l$, kernel size $k$, and $C$ channels, using ReLU activations and residual connections. Regularization includes dropout with probability $p$ after temporal convolutions and $\ell_2$ weight decay of $w_d$. Input features are z-score normalized per channel over the training split. To address class imbalance, inverse-frequency weights are applied to the cross-entropy term. The composite loss uses hyperparameters $\lambda$, $\gamma$, and $\tau$ to balance frame-level accuracy, temporal smoothness, and transition stability.  

Optimization used AdamW with initial learning rate  $\eta_0$, a cosine decay schedule with warmup of $E_w$ epochs, batch size $B$, gradient clipping at $g_{\max}$, and mixed precision. Training proceeds up to $E_{\max}$ epochs with early stopping (patience $P$) monitoring a validation metric; the checkpoint that maximizes a selection metric (F1@50) is reported. At inference, we apply repetition-collapsing and an optional median filter of window $W$ to suppress spurious spikes and improve segment smoothness.

\paragraph{Instantiation for our experiments}
We use $S=4$ stages with $L=10$ layers per stage and \textit{Fibonacci} dilated convolutions; kernel $k=3$, channels $C=64$, ReLU, residual connections; dropout $p{=}0.5$; weight decay $w_d=1\times10^{-4}$. Features are z-score normalized per channel; class imbalance is handled via inverse frequency weights in the cross-entropy term. The loss hyperparameters are $\lambda=0.15$, $\gamma=0.25$, $\tau=4.0$. The models are implemented in PyTorch and trained on a single NVIDIA GTX 1650 (4\,GB VRAM) with AdamW at $\eta_0=5\times10^{-4}$, cosine decay with $E_w=5$ warm-up epochs, batch size $B=8$, gradient clipping $g_{\max}{=}5.0$, and mixed precision. We train up to $E_{\max}{=}50$ epochs with early stopping ($P=5$) on validation loss and report the checkpoint with the best F1@50. In inference, we use repetition collapsing and an optional median filter with $W=3$.

%%%%%%%%%%%%%%%%%%%%%%%%%%%%%%%%%%%%%%%%%%%%%%
% \subsection{Comparison with the State-of-the-Art/ comparative analysis}
\subsection{Comparison with other methods}
The results reported in Table~\ref{tab:comparison_results} compare RoboSubtaskNet with the MS-TCN and MS-TCN++ approaches on GTEA, Breakfast, and RoboSubtask datasets. In GTEA, RoboSubtaskNet outperforms both baselines in boundary-sensitive metrics (F1 @ $10, 25, 50$ = ${91.9, 89.6, 79.5}$ vs. $88.8, 85.7, 76.0$ for MS-TCN++; Edit = ${88.6} \text{ vs.}\ 83.5)$, with comparable frame accuracy $ (78.9 \text{ vs.}\ 80.1)$. In Breakfast, which features longer-horizon activities, our method trails MS-TCN++ across metrics (F1@$50 = 30.4 \text{ vs.}\ 45.9)$, reflecting the trade-off of a denser short-horizon dilation schedule. In the RoboSubtask dataset, where the labels are aligned with the manipulator primitives, RoboSubtaskNet achieves the best scores in all metrics (F1$@50 = {94.2}, \text{ Edit} = {95.6}$,\text{ Accuracy} = ${92.15})$, indicating a strong suitability for robot-executable sub-tasks.

In summary, \textit{RoboSubtaskNet} achieves the best boundary-sensitive metrics on \textit{GTEA} and the best overall scores on \textit{RoboSubtask}, while \textit{MS-TCN++} remains stronger on \textit{Breakfast}, consistent with its long-horizon focus.
\begin{table}[h]
\caption{Comparison with state-of-the-art approaches on GTEA, Breakfast, and the RoboSubtask datasets}
\centering
\resizebox{\columnwidth}{!}{%
\begin{tabular}{|l|c|c|c|c|c|}
\hline
\multicolumn{6}{|c|}{\textbf{GTEA}} \\
\hline
\textbf{Model} & \multicolumn{3}{c|}{\textbf{F1@\{10,25,50\}}} & \textbf{Edit} & \textbf{Acc} \\
\hline
MS-TCN        & 85.8 & 83.4 & 69.8 & 79.0 & 76.3 \\
MS-TCN++      & 88.8 & 85.7 & 76.0 & 83.5 & \textbf{80.1} \\
\textbf{RoboSubtaskNet} & \textbf{91.9} & \textbf{89.6} & \textbf{79.5} & \textbf{88.6} & 78.9 \\
\hline
\multicolumn{6}{|c|}{\textbf{Breakfast}} \\
\hline
\textbf{Model} & \multicolumn{3}{c|}{\textbf{F1@\{10,25,50\}}} & \textbf{Edit} & \textbf{Acc} \\
\hline
MS-TCN        & 52.6 & 48.1 & 37.9 & 61.7 & 66.3 \\
MS-TCN++      & \textbf{64.1} & \textbf{58.6} & \textbf{45.9} & \textbf{65.6} & \textbf{67.6} \\
\textbf{RoboSubtaskNet} & 48.4 & 41.7 & 30.4 & 52.03 & 53.5 \\
\hline
\multicolumn{6}{|c|}{\textbf{RoboSubtask}} \\
\hline
\textbf{Model} & \multicolumn{3}{c|}{\textbf{F1@\{10,25,50\}}} & \textbf{Edit} & \textbf{Acc} \\
\hline
MS-TCN        & 91.5 & 91.4 & 91.0 & 91.8 & 89.2 \\
MS-TCN++      & 92.6 & 92.4 & 92.2 & 93.9 & 91.0 \\
\textbf{RoboSubtaskNet} & \textbf{94.6} & \textbf{94.6} & \textbf{94.2} & \textbf{95.6} & \textbf{92.15} \\
\hline
\end{tabular}}
\label{tab:comparison_results}
\end{table}
%%%%%%%%%%%%%%%%%%%%%%%%%%%%%%%%%%%%%%%%%%%%%%%%%%%%%
\begin{table}[h]
\centering
\setlength{\tabcolsep}{6pt}
\caption{Comparison of different loss functions on the RoboSubtask dataset with the proposed approach}
\label{tab:loss_RoboSubtask}
\begin{tabular}{|l|c|c|c|c|c|}
\hline
\textbf{Loss} & \multicolumn{3}{c|}{\textbf{F1@\{10, 25, 50\}}} & \textbf{Edit} & \textbf{Acc} \\
\hline
${L}_{\mathrm{CE}}$ & 92.3 & 91.9 & 90.1 & 90.4 & 89.8 \\
\hline
${L}_{\mathrm{CE}}+\lambda\,{L}_{\mathrm{T\text{-}MSE}}$ & 92.9 & 92.0 & 91.5 & 91.5 & 90.6 \\
\hline
${L}_{\mathrm{CE}} + \lambda\, {L}_{\mathrm{T\text{-}MSE}} + \gamma\, {L}_{\mathrm{Trans}}$ & \textbf{94.6} & \textbf{94.6} & \textbf{94.2} & \textbf{95.6} & \textbf{92.15} \\
\hline
\end{tabular}
\end{table}
%%%%%%%%%%%%%%%%%%%%%%%%%%%%%%%%%%%%%%%%%%%%%%%%%%%%%
\begin{figure*}[ht]
    \centering

    % ================== Dataset 1: GTEA ==================
    \begin{subfigure}[b]{0.47\textwidth}
        \centering
        \fbox{\includegraphics[width=1\linewidth]{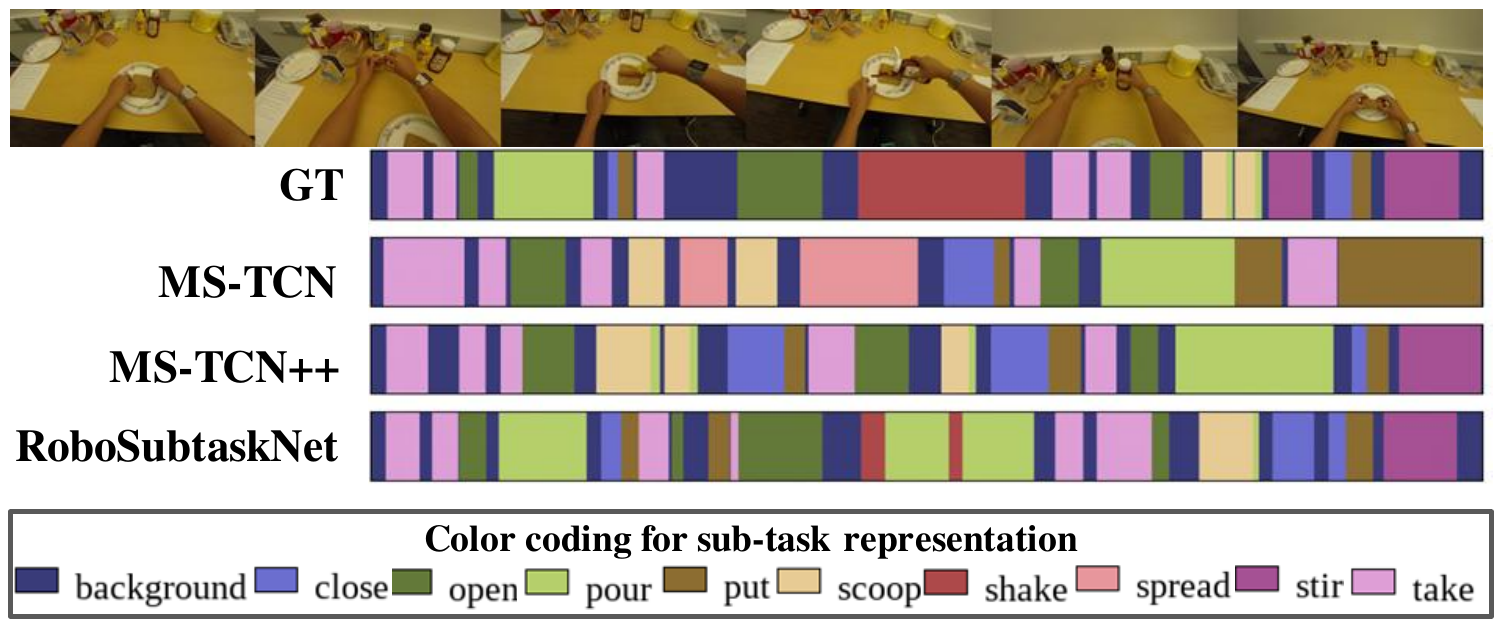}}
        \caption{}
    \end{subfigure}
    \hfil
    \begin{subfigure}[b]{0.47\textwidth}
        \centering
        \fbox{\includegraphics[width=1\linewidth]{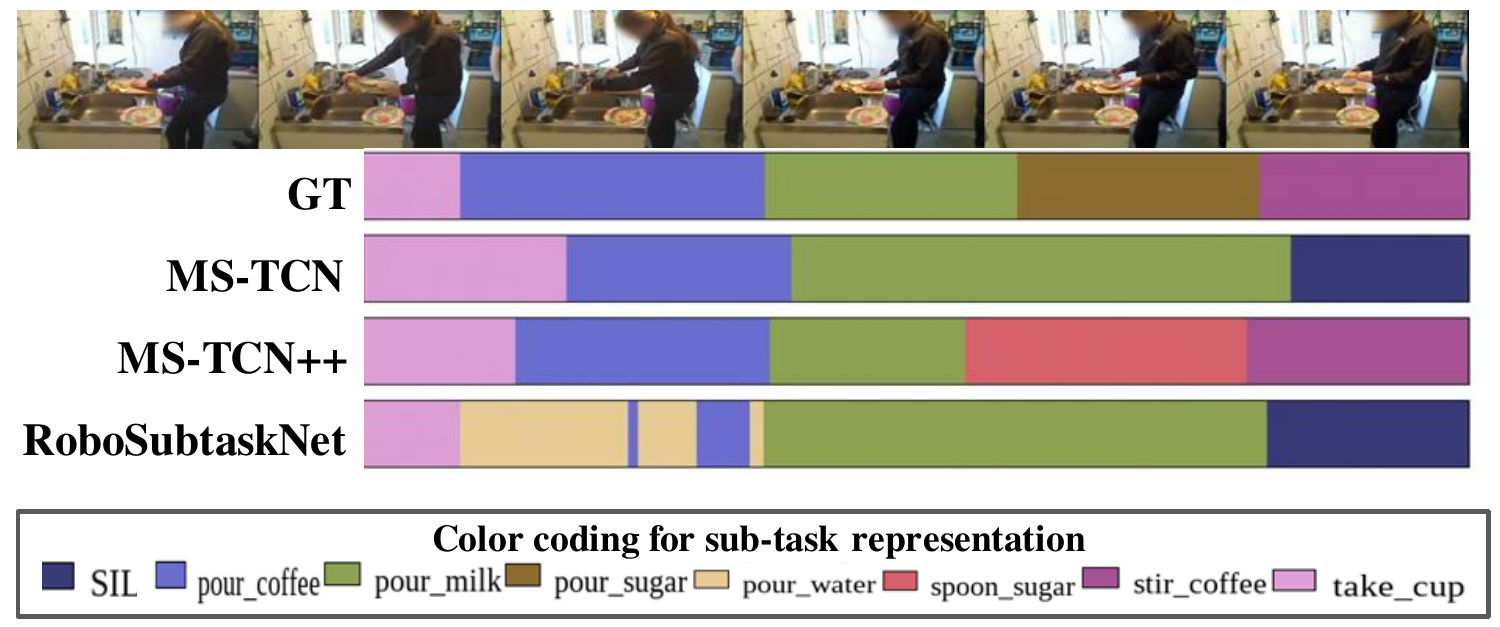}}
        \caption{}
    \end{subfigure}
    
    \begin{subfigure}[b]{0.47\textwidth}
        \centering
\fbox{\includegraphics[width=1\linewidth]{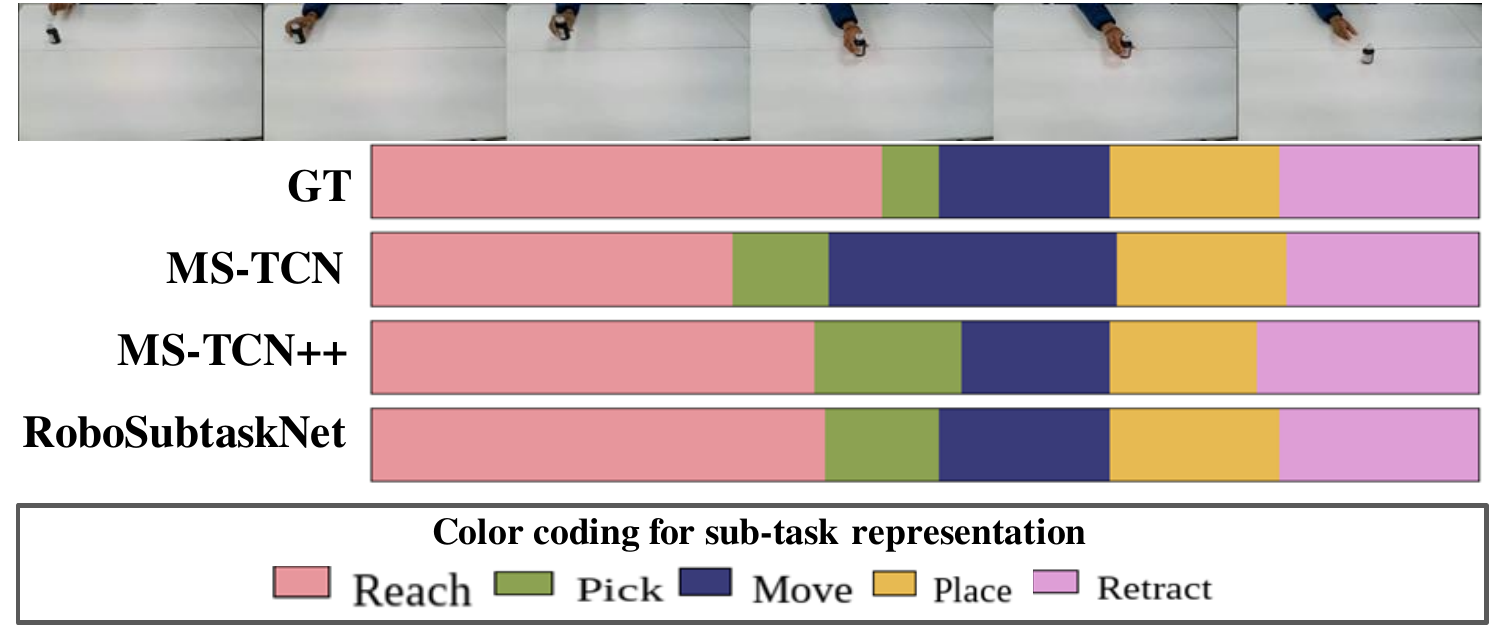}}
        \caption{}
    \end{subfigure}
    \hfil
    \begin{subfigure}[b]{0.47\textwidth}
        \centering
        \fbox{\includegraphics[width=1\linewidth]{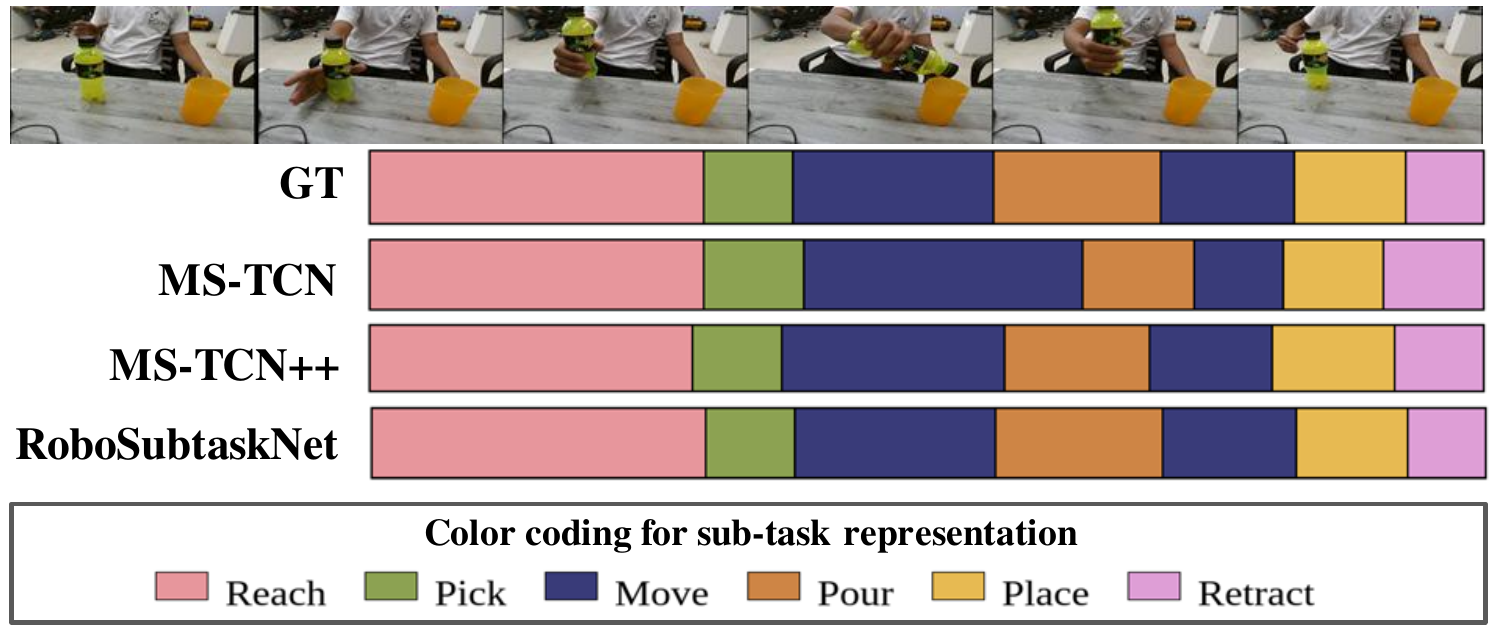}}
        \caption{}
    \end{subfigure}
    \caption{Qualitative results of RoboSubtaskNet with MS-TCN and MS-TCN++ approaches. One example video is shown from each dataset: (a) GTEA and (b) Breakfast. Two examples (c), (d) from RoboSubtask for \textit{pick and place} and \textit{pick and pour} tasks, respectively. For each example, the video collage is presented along with the ground truth and predicted segmentation.}
\label{fig:qualitative_results_all}
\end{figure*}

\subsection{Ablations and qualitative analysis}
Table~\ref{tab:loss_RoboSubtask} summarizes experiments that isolate the effect of different loss combinations on the RoboSubtask dataset. Using only the cross-entropy loss yields reasonable baseline performance. Augmenting cross-entropy with the temporal smoothing term (T-MSE) produces a small improvement in the Edit score from $64.2$ to $64.6$. In contrast, adding the transition-aware loss together with T-MSE yields large improvements across all metrics: F1@50 increases from $60.7$ to $98.0$, Edit increases from $64.2$ to $98.5$, and frame accuracy increases from $79.9$ to $96.2$. These results indicate that explicitly penalizing rapid probability changes and implausible label switches is especially important for accurate segmentation of short-horizon, robot-executable sub-tasks.

Figure~\ref{fig:qualitative_results_all} presents qualitative results: one video from each GTEA and Breakfast, and two videos from RoboSubtask (pick and place and pick and pour). For each example, we display the video collage, ground-truth segmentation, and predictions from MS-TCN, MS-TCN++, and RoboSubtaskNet. On GTEA and RoboSubtask, RoboSubtaskNet produces cleaner boundaries and fewer over-segmentation artifacts than the baselines, particularly around brief transitions such as reach to pick and place to retract. In Breakfast, which contains long-horizon composite activities, RoboSubtaskNet remains competitive with MS-TCN but occasionally under-segments extended activities, consistent with the method’s design emphasis on dense local coverage. These qualitative and quantitative outcomes can be attributed to the combination of attention-based fusion which balances RGB and flow cues, the Fibonacci dilation schedule which provides denser short-horizon temporal coverage, and the transition-aware loss which suppresses implausible label switches, together yielding robust sub-task segmentation for robot-executable demonstrations.
%%%%%%%%%%%%%%%%%%%%%%%%%%%%%%%%%%%%%%%%%%%%%%%%%%%%%%%%%%%%%%%%%%%%%%%%%%%%%%%%%%%%%%%%%%%%%%%%%%%%%%%%%%%%%%%%%%%%%%%%%%%%%%%%%%%%%%%%%%
\subsection{Real-world tasks evaluation}
We evaluated the overall performance of our methodology on four representative manipulation tasks. Figure~\ref{fig:pick_place} illustrates the pick and place task, Figure~\ref{fig:pick_pour} depicts the pick and pour task, Figure~\ref{fig:clean_table} shows the cleaning task of the work surface, and Figure~\ref{fig:pick_give} presents the pick and give task. Each figure demonstrates the robot's ability to learn from human demonstrations, identify the individual sub-tasks performed in those demonstrations, and execute the appropriate sub-task sequence using the robot execution model.

\subsubsection{Task performance and execution analysis}
We evaluated the proposed system on four distinct manipulation tasks, conducting $20$ tests for each. The results, summarized in Table~\ref{tab:combined-eval}, demonstrate the robustness of our approach. The system achieved $95\%$ success rate for both table cleaning $\&$ pick and place tasks, and $90\%$ for pick and pour task. The interactive pick and give task produced an $85$\% success rate.  

An important aspect of our evaluation is the time needed for the prediction of sub-tasks from video demonstrations. In our experiments, the input videos ranged from $3$ to $25$ seconds (approximately $90$ to $750$ frames at $30$ fps). Shorter inputs ($3$ seconds or $90$ frames) required less than $10-15$ seconds for sub-task prediction, while the longest inputs ($25$ seconds or $750$ frames) took about $1-1.5$ minutes. This scalability supports the feasibility of our method for real-time applications.

\begin{table}[!ht]
  \centering
  \caption{Overall task-wise evaluation}
  \resizebox{\columnwidth}{!}{%
    \begin{tabular}{|l|c|c|c|c|c|}
      \hline
      Task & \makecell{Successful \\ runs} & \makecell{Failed \\ runs} 
           & \makecell{Success \\ rate (\%)} & \makecell{Sub-task \\ prediction time (s)} 
           & \makecell{Task execution \\ time (s)} \\
      \hline
      Pick \& Pour   & 18 & 2 & 90 & 60--90 & 50--60 \\
      Table Cleaning & 19 & 1 & 95 & 60--90 & 50--60 \\
      Pick \& Give   & 17 & 3 & 85 & 60--90 & 30--40 \\
      Pick \& Place  & 19 & 1 & 95 & 60--90 & 45--50 \\
      \hline
    \end{tabular}
    }
    \label{tab:combined-eval}
\end{table}

For physical execution, the robot completed the tasks in $30$ to $60$ seconds, with variations depending on the complexity and duration of the sub-tasks required for each task.

\subsubsection{Failure analysis}
In all $80$ trials, our system achieved a $91.25\%$ success rate. We analyze $7$ failures in all these tasks. As detailed in Table~\ref{tab:failure_analysis}, issues occurred at the execution level: $42.85\%$ from the physical robot execution model; these failures were caused by perception errors such as object detection inaccuracy, motion singularities, and grasp slippage, the other $28.57\%$ from general system failures (e.g., system lags, robot hangs). The remaining $28.57\%$ of failures originated in the planning stage, caused by an incorrect prediction of sub-task sequence.

\begin{table}[H]
    \centering
    \caption{Failure analysis (7 failures across all tasks)}
    \label{tab:failure_analysis}
    \begin{tabular}{|l|c|c|}
      % \hline
      % \multicolumn{3}{|c|}{\textbf{Failure Analysis}} \\
      \hline
      Failure Cause                   & \ Failures & Percentage (\%) \\
      \hline
      Robot execution model            & 3           & 42.85             \\
      Sub-task Sequence Prediction   & 2           & 28.57               \\
      System Failure                  & 2           & 28.57             \\
      \hline
    \end{tabular}
  \end{table}
%%%%%%%%%%%%%%%%%%%%%%%%%%%%%%%%%%%%%%%%%%%%%%%%%%%%%%%%%%%%%%
\section{Conclusion}
We presented a multi-stage \emph{human-to-robot sub-task segmentation} framework that couples attention-enhanced I3D features (RGB+Flow) with a modified MS-TCN using a Fibonacci dilation schedule and transition-aware regularization. To bridge vision benchmarks and robotic execution, we introduced RoboSubtask, a sub-task annotated dataset in healthcare and industrial settings, and validated end-to-end deployment on a 7-DoF Kinova Gen3 by deterministically mapping predicted sub-task labels to primitive parameters and executing the resulting sequence. The experiments further show strong segmentation performance on the datasets (GTEA, RoboSubtask) across segmental F1@$\{10,25,50\}$ and Edit score, underscoring that the proposed perception stack improves both recognition quality and downstream robot reliability.\par
Our current system relies on supervised sub-task labels, short-horizon manipulation assumptions, and monocular RGB+flow sensing; mapping from labels to skills is deterministic and only lightly monitored at run-time. Future work will explore (i) self-supervised and active annotation strategies to scale sub-task labeling, (ii) tighter perception–control coupling via closed-loop policies with uncertainty and safety-aware planning, and (iii) longer-horizon task composition segmentation and failure recovery. We believe these directions will further strengthen the bridge from fine-grained video understanding to robust, deployable human–robot collaboration.

%%%%%%%%%%%%%%%%%%%%%%%%%%%%%%%%%%%%%%%%%%%%%%%%%%%%%%%%
\bibliographystyle{ieeetr}
\bibliography{ref}

\end{document}